\documentclass[10pt, a4paper]{article}

\usepackage {lrec2026} 
\usepackage{tikz}
\usetikzlibrary{positioning,arrows.meta,calc}
\usepackage{graphicx} 
\usepackage{linguex}
\usepackage[table]{xcolor}
\usepackage{booktabs}
\usepackage{enumitem}
\usepackage[most]{tcolorbox}
\usepackage{placeins}
\usepackage{pgfplots}
\usepackage{hyperref}
\usepackage{tabularx}
\usepackage{amssymb}

\pgfplotsset{compat=1.18} 
\newtcblisting{PromptBox}[1]{%
  breakable,
  colback=gray!5,
  colframe=black!40,
  title={#1},
  listing only,
  listing options={
    basicstyle=\ttfamily\footnotesize,
    breaklines=true,
    breakatwhitespace=true
  }
}

\definecolor{EviRoot}{RGB}{232,240,254}   
\definecolor{EviDirect}{RGB}{232,245,233} 
\definecolor{EviIndirect}{RGB}{255,243,224} 
\definecolor{EviSubA}{RGB}{243,248,255}   
\definecolor{EviSubB}{RGB}{255,249,240}   
\definecolor{EviNote}{RGB}{245,245,245}   

\title{Transfer Learning for an Endangered Slavic Variety: Dependency Parsing in Pomak Across Contact-Shaped Dialects}

\name{Sercan Karaka{\c{s}}}

\address{University of Chicago \\
         skarakas@uchicago.edu\\}

\abstract{
This paper presents new resources and baselines for Dependency Parsing in Pomak, an endangered Eastern South Slavic language with substantial dialectal variation and no widely adopted standard. We focus on the variety spoken in Turkey (Uzunköprü) and ask how well a dependency parser trained on the existing Pomak Universal Dependencies treebank, which was built primarily from the variety that is spoken in Greece, transfers across dialects. We run two experimental phases. First, we train a parser on the Greek-variety UD data and evaluate zero-shot transfer to Turkish-variety Pomak, quantifying the impact of phonological and morphosyntactic differences. Second, we introduce a new manually annotated Turkish-variety Pomak corpus of 650 sentences and show that, despite its small size, targeted fine-tuning substantially improves accuracy; performance is further boosted by cross-variety transfer learning that combines the two dialects.
\\ \newline \Keywords{Pomak, Universal Dependencies, dialectal variation, cross-lingual transfer, low-resource NLP}
}

\begin{document}
\pagestyle{plain} 

\maketitleabstract
\footnotetext[1]{Pomak also exhibits dialectal variation within Turkey. This paper is based on the dialect spoken in Uzunköprü, Edirne.}
\setcounter{footnote}{1}
\section{Introduction}

Pomak is an endangered Eastern South Slavic language spoken in Bulgaria, Greece, and Turkey, with additional diaspora communities \citep{constantinides2007,kehaya2017,karakas-2020-low-high-applicatives-pomak,karakas2022,adamou2022,markantonatou2023}. Sociolinguistic syntheses and recent NLP-facing surveys characterize Pomak as severely constrained along multiple vitality dimensions, most prominently intergenerational transmission, functional domain coverage, and institutional/community support dimensions, which closely track UNESCO-style endangerment frameworks \citep{brenzinger2003,markantonatou2023}. From an NLP and resource-construction perspective, Pomak poses a compound low-resource problem: (i) the absence of a widely adopted orthographic standard limits the availability and comparability of written data; and (ii) the speech community forms a dialect continuum whose structural properties have been reshaped by sustained contact with Greek and Turkish in settings of pervasive bi-/trilingualism \citep{kehaya2017,karakas2022}. 

More generally, the kinds of lexically and structurally mediated differences observed across Pomak varieties are consistent with well-studied pathways of contact-induced change, in which sustained bilingualism can trigger borrowing, calquing, and convergence effects that extend beyond the lexicon into morphosyntax and argument-structure realizations. Classic work argues that contact outcomes depend on both social factors (intensity, community structure, language attitudes) and linguistic constraints, and that structural transfer can be systematic even in the absence of full language shift \citep{thomasonKaufman1988,winford2003}. In addition, grammatical replication accounts emphasize that contact may promote the diffusion of abstract patterns (e.g., functional-marking strategies) rather than direct morpheme import, yielding typologically coherent but variety-specific reorganizations of form--function mappings \citep{heineKuteva2005,matras2009}. This perspective motivates treating the Turkey variety not merely as a noisier or less standardized sample of ``the same'' Pomak, but as a dialect shaped by identifiable contact pressures whose effects are linguistically structured rather than random. On this view, cross-variety differences should not be reduced to superficial lexical substitution or orthographic inconsistency alone, but understood as reflecting contact-conditioned reorganization in the mapping between form and function. This has direct consequences for computational modeling, because it means that cross-variety generalization may fail not simply due to data sparsity, but because the parser is exposed to systematically different distributions of cues in the target variety. It also implies that robust parsing may require a greater amount of in-variety supervision than would be expected if the two varieties differed only minimally. Consequently, variation is not merely lexical but systematically grammatical, affecting phonological inventories, morphosyntactic exponence, and argument-realization options. To give an example, prior descriptive and comparative work documents dialect-sensitive differences in nominal case morphology and oblique marking strategies \citep{sandry2013}, and shows that closely related varieties can diverge in the syntax--morphology mapping for ditransitives (e.g., the encoding of goal/recipient arguments via inflectional case vs.\ marker/adpositional strategies) \citep{karakas2022}. The contrast in the following sentences exemplifies such micro-variation by comparing the Xanthi (Greece) and Uzunköprü (Turkey) varieties, highlighting consequences for cross-variety generalization in supervised tagging and transfer-based modeling.

\ex.\label{ex:pomak-ditrans}

\noindent(a) \textit{Xanthi (Greece)}

\gll Ayşe dade resima Aliyu. \\
     Ayşe give.\textsc{pst.3sg} picture-\textsc{def} Ali-\textsc{dat} \\
\trans `Ayşe gave Ali the picture.'

\medskip

\noindent(b) \textit{Uzunköprü (Turkey)}

\gll Ayşe resim-asa na Ali dade. \\
     Ayşe picture-\textsc{poss}.\textsc{def} \textsc{dat.mrk} Ali give.\textsc{pst.3sg} \\
\trans `Ayşe gave Ali his/her picture.' \footnotemark
\footnotetext{Depending on the variety and analysis, \textit{na} has been treated either as a preposition-like element or as a dedicated marker introducing goal/recipient arguments; we gloss it neutrally as \textsc{dat.mrk} here.}

As the example shows, the Xanthi variety licenses an inflectional dative on the recipient (Ali-\textsc{dat}), whereas the Uzunköprü variety introduces the recipient with \textit{na}, with no overt dative suffix on the proper name. This alternation is consistent with broader claims that Pomak varieties differ in how they encode oblique arguments and distribute case marking across nominal morphology vs.\ adpositional/marker strategies \citep{sandry2013,karakas2022}.

Given the very limited resources available for Pomak, especially for the variety spoken in Turkey, the study is structured in two phases aimed at developing computational tools to support its processing. The first phase involves building a dependency parser for the Pomak variety spoken in Turkey. This effort builds on the linguistic description provided by \citet{karakas2022} for the Turkey variety and on existing morphologically annotated corpora and UD-style resources developed for the Greece variety \citep{jusuf-karahoga-etal-2022-morphologically, markantonatou2023}. The central research question is whether a parser trained on the Greece variety can effectively accommodate linguistic variation in the Turkey variety, given that Pomak has been documented to show dialectal variation in prior work \citep{sandry2013,kehaya2017,adamou2018,adamou2022,karakas2022}.

The second phase of this study introduces a newly developed corpus based on the Pomak variety spoken in Turkey. This corpus serves to evaluate whether the parser can effectively be trained even on a small corpus of the targeted dialect. Due to the limited size of the available data, which results in a poor performance, cross-lingual transfer learning techniques were employed as a solution to overcome the data scarcity. The results demonstrate that cross-lingual transfer learning significantly improves the model’s performance, providing the highest accuracy scores when applied to the Pomak dialect which is spoken in Turkey. By addressing these challenges, this study contributes to the growing body of research focused on enhancing computational resources for low-resourced and dialectally diverse languages, with particular emphasis on Pomak.

\section{Related Work}
Foundational work on Pomak has established both (i) detailed dialect-specific grammatical descriptions and (ii) a growing set of corpus-linked, computationally usable linguistic resources. For the Paševik (Greece) variety, \citet{sandry2013} provides a broad-coverage description of phonology and nominal/verbal morphology with accompanying notes on core syntax and annotated texts, offering one of the most systematic documentary baselines for Pomak micro-variation. In parallel, generative analyses have shown that dialectal differences are not merely lexical, but can reflect deeper differences in argument structure and syntax--semantics mapping: for example, \citet{karakas2022} argues (on the basis of binding, scope, and weak crossover diagnostics) that the Xanthi (Greece) variety patterns with low-applicative double-object structures, whereas the Uzunköprü (Turkey) variety instead aligns with prepositional-ditransitive configurations. 

Beyond structural description, recent sociolinguistic work has also clarified how contact ecologies and speaker attitudes bear directly on maintenance. In particular, \citet{dagdevirenKirmizi2026insecurity} investigates linguistic (in)security among Pomak heritage speakers in Western Türkiye via semi-structured interviews, reporting systematic age- and language-contingent proficiency profiles and pervasive insecurity across interactional settings. Importantly for resource-building, the study links family language practices and failures of intergenerational transmission to experienced (in)security and to subtractive schooling effects in the dominant language, which may show that data scarcity is partly socially produced rather than purely technological.

In parallel, there is rapidly growing NLP-oriented interest in Pomak, with recent work targeting the creation of reusable digital infrastructures and open resources \citet{krimpas-etal-2021-pomak}. On the text side, \citet{jusuf-karahoga-etal-2022-morphologically} describe the development of a Latin-based orthography and gold morphological annotation for Pomak, explicitly aligning the resulting resources with UD-style dissemination and downstream tool training. Complementing this, \citet{kokkas2025exploration} report the digitization and exploration of a Pomak corpus compiled in the \emph{Philotis} project (2020--2023), comprising roughly 140k words across diverse genres and released as an open textual resource intended for NLP practitioners and quantitative corpus analysis. Extending beyond text, \citet{tsoukala-etal-2023-asr} present an end-to-end pipeline for training ASR systems for low-resourced languages and use Pomak as a case study, training what they describe as the first Pomak ASR model. These efforts motivate a modeling agenda in which cross-variety transfer, orthography/annotation decisions, and modality-specific pipelines are treated as first-class design variables for Pomak NLP.

\section{Methods}

To assess whether Pomak’s documented dialectal heterogeneity translates into measurable degradation in syntactic generalization, we fine-tuned a state-of-the-art graph-based neural dependency parser \citep{dozat2017} on the Pomak Universal Dependencies (UD) treebank \citep{jusuf-karahoga-etal-2022-morphologically}. The treebank is derived from the QPM corpus with syntactic annotation guided by an active-learning workflow \citep{settles2009,anastasopoulos2018,shi2021}. Concretely, the active annotation strategy iteratively prioritizes informative sentences for manual correction, thereby concentrating scarce expert effort on examples expected to yield the largest model and guideline improvements, which can be considered to be a particularly appropriate design choice for a non-standardized, low-resource language with substantial intra-variety variation.

All parsing experiments use our own implementation of the biaffine, graph-based dependency parser of \citet{dozat2017}. The reason why we adopt a graph-based parser is because it directly optimizes over full dependency graphs and has been shown to be robust for morphologically rich languages, where fusional exponence and flexible word order increase the density of long-distance dependencies and raise ambiguity for locally greedy transition systems \citep{dozat2017}. This architectural bias is especially relevant for Pomak, whose Slavic-type morphology and contact-shaped micro-variation can interact to produce non-trivial argument-structure and case-marking patterns. 


\subsection{Parser Architecture}\label{sec:parser}

We implement the biaffine, graph-based dependency parser of \citet{dozat2017}. Given a sentence of tokens $w_{1:n}$, the model (i) builds contextual token representations, (ii) scores all possible head--dependent arcs with a biaffine scorer, (iii) scores dependency relation labels for each chosen arc, and (iv) decodes the highest-scoring well-formed dependency tree (non-projective maximum spanning arborescence) and assigns labels.

\paragraph{Token representations and contextual encoder.}
Each token $w_i$ is represented by a concatenation of a word embedding and a
character-composed embedding (optionally augmented by discrete features):
\ex.\label{ex:parser-repr}
\a. \textbf{Input representation.}
\[
\begin{aligned}
x_i &= \big[e^{(w)}_i;\; e^{(c)}_i;\; e^{(\mathrm{feat})}_i\big], \\
e^{(c)}_i &= \mathrm{CharEnc}(\mathrm{ch}_{i,1:m_i}).
\end{aligned}
\]
\b. \textbf{Contextualization.}
\[
r_i = \mathrm{BiLSTM}(x_{1:n})_i .
\]

\paragraph{Arc scoring (biaffine).}
We produce separate head and dependent projections for arc prediction using MLPs, then score every candidate head $j$ for dependent $i$:
\ex.\label{ex:parser-arc}
\a. \textbf{Arc-specific projections.}
\[
d_i^{a}=\mathrm{MLP}^{\mathrm{dep},a}(r_i),
\qquad
h_j^{a}=\mathrm{MLP}^{\mathrm{head},a}(r_j).
\]
\b. \textbf{Biaffine arc score.}
\[
s^{a}_{ij}
=
(d_i^{a})^{\top} U^{a} h_j^{a}
+
(u^{a})^{\top}\!\big[d_i^{a};h_j^{a}\big]
+
b^{a}.
\]
\c. \textbf{Head distribution.}
\[
p(h_i=j\mid i)=\mathrm{softmax}_{j}\!\left(s^{a}_{ij}\right),
\]
where $h_i$ denotes the head index of token $i$ (including a designated \textsc{root}).

\paragraph{Relation labeling (biaffine).}
Conditioned on a head choice $j$ for dependent $i$, we score a distribution over relation labels $\ell\in\mathcal{R}$:
\ex.\label{ex:parser-rel}
\a. \textbf{Relation-specific projections.}
\[
d_i^{r}=\mathrm{MLP}^{\mathrm{dep},r}(r_i),
\qquad
h_j^{r}=\mathrm{MLP}^{\mathrm{head},r}(r_j).
\]
\b. \textbf{Biaffine relation score (vector over labels).}
\[
s^{r}_{ij}
=
(d_i^{r})^{\top} U^{r} h_j^{r}
+
W^{r}\!\big[d_i^{r};h_j^{r}\big]
+
b^{r},
\]
where $s^{r}_{ij}\in\mathbb{R}^{|\mathcal{R}|}$, $U^{r}$ is a label-tensor, and
$W^{r},b^{r}$ are label-wise affine parameters.
\c. \textbf{Relation distribution.}
\[
p(r_i=\ell \mid i, h_i=j)=\mathrm{softmax}_{\ell}\!\left(s^{r}_{ij}[\ell]\right),
\]
where $r_i$ denotes the relation label of token $i$.

\paragraph{Decoding.}
At inference time, we decode a single well-formed dependency tree using the arc scores
$\{s^{a}_{ij}\}$ (allowing non-projective structures) and then assign a label to each
selected arc:
\ex.\label{ex:parser-decode}
\a. \textbf{Tree decoding (arcs).}
\[
\hat T
=
\arg\max_{T\in\mathcal{T}}
\sum_{(j\rightarrow i)\in T} s^{a}_{ij},
\]
where $\mathcal{T}$ is the set of valid dependency trees (arborescences).
\b. \textbf{Label assignment.}
\[
\hat r_i=\arg\max_{\ell\in\mathcal{R}} s^{r}_{i\,\hat h_i}[\ell],
\]
with $\hat h_i$ given by the decoded tree $\hat T$.

\paragraph{Training objective.}
Let $j_i^{\star}$ be the gold head of token $i$ and $\ell_i^{\star}$ the gold relation label.
We minimize the negative log-likelihood of (i) head selection and (ii) relation labeling:
\ex.\label{ex:parser-loss}
\a. \textbf{Arc loss.}
\[
\mathcal{L}_{\text{arc}}
= -\sum_{i=1}^{n}\log p\!\left(h_i=j_i^{\star}\mid i\right)
\]
\b. \textbf{Relation loss.}
\[
\mathcal{L}_{\text{rel}}
= -\sum_{i=1}^{n}\log p\!\left(r_i=\ell_i^{\star}\mid i, h_i=j_i^{\star}\right)
\]
\c. \textbf{Total objective.}
\[
\mathcal{L}
=
\mathcal{L}_{\text{arc}}+\mathcal{L}_{\text{rel}}+\lambda\|\theta\|_2^2 .
\]

This formulation cleanly separates head selection (UAS-relevant) from relation labeling (LAS-relevant), matching the evaluation in Section~\ref{sec:results}.
\subsection{A New Corpus}\label{sec:new-corpus}


\definecolor{BoxDataA}{RGB}{232,240,254}   
\definecolor{BoxDataB}{RGB}{232,245,233}   
\definecolor{BoxModel}{RGB}{255,243,224}   
\definecolor{BoxEval}{RGB}{245,245,245}    
\definecolor{EdgeA}{RGB}{46,92,150}
\definecolor{EdgeB}{RGB}{56,120,88}
\definecolor{EdgeX}{RGB}{170,110,40}

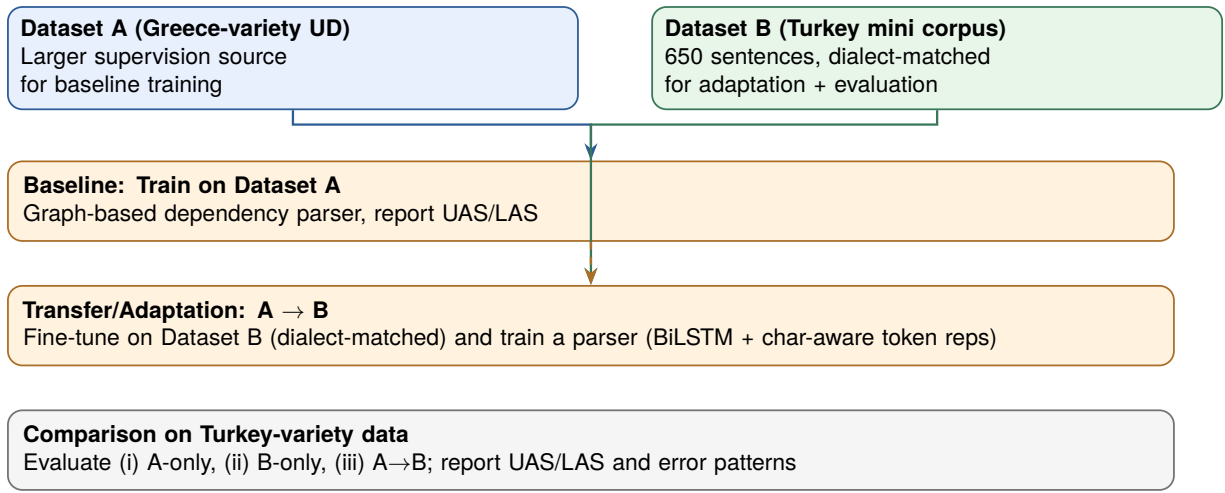
\begin{figure*}[!t]
\centering
\resizebox{\textwidth}{!}{%
\begin{tikzpicture}[
  font=\small,
  >={Stealth[length=2.4mm]},
  smallbox/.style={draw, rounded corners=2mm, line width=0.7pt,
                   align=left, inner sep=5pt, text width=0.47\textwidth},
  bigbox/.style={draw, rounded corners=2mm, line width=0.7pt,
                 align=left, inner sep=6pt, text width=0.98\textwidth},
  a/.style={smallbox, fill=BoxDataA, draw=EdgeA},
  b/.style={smallbox, fill=BoxDataB, draw=EdgeB},
  m/.style={bigbox, fill=BoxModel, draw=EdgeX},
  e/.style={bigbox, fill=BoxEval, draw=black!55},
  arrA/.style={->, draw=EdgeA, line width=0.9pt},
  arrB/.style={->, draw=EdgeB, line width=0.9pt},
  arrX/.style={->, dashed, draw=EdgeX, line width=0.9pt}
]

\node[a] (Adata) {\textbf{Dataset A (Greece-variety UD)}\\
Larger supervision source\\
for baseline training};
\node[b, right=10mm of Adata] (Bdata) {\textbf{Dataset B (Turkey mini corpus)}\\
650 sentences, dialect-matched\\
for adaptation + evaluation};

\node[m, below=7mm of Adata.south west, anchor=north west] (baseline)
{\textbf{Baseline: Train on Dataset A}\\
Graph-based dependency parser, report UAS/LAS};

\node[m, below=6mm of baseline.south west, anchor=north west] (adapt)
{\textbf{Transfer/Adaptation: A $\rightarrow$ B}\\
Fine-tune on Dataset B (dialect-matched) and train a parser (BiLSTM + char-aware token reps)};

\node[e, below=6mm of adapt.south west, anchor=north west] (eval)
{\textbf{Comparison on Turkey-variety data}\\
Evaluate (i) A-only, (ii) B-only, (iii) A$\rightarrow$B; report UAS/LAS and error patterns};

\draw[arrA] (Adata.south) -- ++(0,-2mm) -| (baseline.north);
\draw[arrB] (Bdata.south) -- ++(0,-2mm) -| (adapt.north);
\draw[arrX] (baseline.south) -- (adapt.north);

\end{tikzpicture}%
}
\vspace{-2mm}
\caption{Workflow summary. Dataset A provides baseline supervision; Dataset B provides dialect-matched data for adaptation and evaluation.}
\label{fig:methods-overview}
\vspace{-2mm}
\end{figure*}

Recall that the second phase of the study addresses a central limitation of existing Pomak NLP resources: the lack of dialect-matched annotated data for the variety spoken in Turkey. Although the currently available Pomak UD treebank has made initial modeling possible, it mainly reflects the Greece variety and thus can be said to capture lexical choices, morphological patterns, and syntactic distributions that are not always representative of the Uzunköprü variety. This matters because dialectal differences are not limited to pronunciation or a small number of vocabulary items as we established in the previous sections. They can also affect the forms of case marking, the realization of verbal morphology, the choice between competing function words or adpositions, and the distribution of dependency relations in context. In the Uzunköprü variety, these contrasts are further shaped by sustained contact with Turkish, which may influence both lexical selection and morphosyntactic preferences. As a result, a model trained only on the Greece-variety treebank may learn analyses that are well suited to that variety but systematically mis-handle tokens and structures that are natural in Uzunköprü. Creating dialect-matched annotated data is therefore necessary not only for improving accuracy, but also for testing how well current NLP methods transfer across closely related yet contact-shaped Pomak varieties. This is particularly consequential in Pomak, where sustained bilingualism and long-term contact with Turkish can yield contact-induced divergence in both the lexicon and the grammar, affecting the surface realization of functional material and argument-marking strategies \citep{karakas2022,thomasonKaufman1988,winford2003,heineKuteva2005,matras2009}. In low-resource settings, such distributional shifts are amplified: even small differences in case/marker realizations, cliticization patterns, or function-word inventories can systematically alter the feature space that a tagger or parser relies on, thereby degrading cross-variety transfer.

To enable controlled evaluation and adaptation, we compiled a mini corpus of the Pomak variety spoken in Turkey, consisting of 650 sentences drawn from two complementary sources. First, we collected sentence-level data from six native speakers of the Turkey variety of Pomak living in Turkey. These speakers were between 47 and 60 years old and were bilingual in Pomak and Turkish. The speaker-contributed portion of the corpus primarily reflects everyday language use rather than highly formal, literary, or specialized registers. Second, we incorporated additional sentences from a study that provides primary data on the Turkey variety \citep{karakas2022}. Therefore, the corpus should be understood as a small mixed-source sentence-level resource rather than a genre-balanced corpus.

Because Pomak lacks a widely adopted orthographic standard, corpus preparation involved an explicit normalization and annotation workflow. We first normalized orthographic forms and standardized tokenization so that the data could be used consistently for supervised modeling and compared directly with existing UD-style Pomak resources. We then annotated the corpus in accordance with the Universal Dependencies framework, assigning POS tags, morphological features, head--dependent relations, and dependency labels in a UD-compatible format. This workflow was designed to ensure maximal comparability between the existing Greece-variety UD resource and the newly compiled Turkey-variety corpus, while also making the new dataset suitable for transfer-based parsing experiments.

Despite its usefulness, the corpus size is intentionally small and therefore represents the ``extreme low-resource'' regime that characterizes endangered-language NLP in practice. When we train the syntactic pipeline exclusively on this 650-sentence corpus, performance is substantially lower than models trained on the larger Greece-variety UD resource, reaching only 50 UAS and 44 LAS. This result is informative in two respects. First, it quantifies the limits of purely within-variety supervision at this scale: even with dialect-matched data, the model cannot learn sufficiently robust generalizations from 650 sentences. Second, it motivates the transfer-learning strategy pursued in later sections, where cross-variety supervision from the larger Pomak UD resource is combined with targeted fine-tuning on the Turkey-variety mini corpus. In other words, the mini corpus is not intended to replace existing resources, but to serve as a crucial \emph{adaptation set} that anchors the model in the distributional and contact-shaped properties of the Turkey variety while leveraging larger out-of-variety supervision for statistical strength.

We distinguish three datasets. Dataset A is the existing Pomak UD treebank
(primarily Greece variety). Dataset B is our newly compiled Turkey-variety corpus
of 650 sentences, drawn from bilingual Pomak--Turkish speakers living in Turkey and from prior descriptive materials, which we split 80/20 into 520 sentences for training and 130
sentences for development (early stopping and model selection). Finally, Dataset C is a
held-out Turkey-variety test set of 90 sentences collected from a \emph{different native
speaker} than those contributing to Dataset~B (speaker-disjoint evaluation). All reported
Turkey-variety results (Tables~\ref{tab:uas_las}--\ref{tab:scores_all}) are computed on
Dataset~C, ensuring comparability across training regimes.

\section{Evaluation Metrics and Results}\label{sec:results}

We evaluate syntactic parsing performance using two standard dependency-parsing metrics: unlabeled attachment score (UAS) and labeled attachment score (LAS). UAS measures the proportion of tokens whose predicted head matches the gold head, abstracting away from relation type. LAS is stricter: a token is counted as correct only if \emph{both} the predicted head and the dependency relation label match the gold annotation. Throughout, scores are computed at the word level and exclude punctuation following common UD evaluation conventions.

Table~\ref{tab:uas_las} reports baseline cross-variety performance when the parser is trained on the Greece-variety Pomak UD resource and evaluated on Turkey-variety Pomak. The baseline achieves 59\% UAS and 51\% LAS. Two aspects of this pattern are noteworthy. First, the gap between UAS and LAS (8 points) indicates that a non-trivial share of errors involve relation labeling \emph{in addition to} head selection, consistent with the possibility that dialect-conditioned differences affect not only local attachment preferences but also the distribution of functional material and morphosyntactic cues used to disambiguate relation types. Second, the overall level of performance is substantially below what is typically observed when training and testing are conducted within the same variety under comparable annotation conditions, suggesting that cross-variety transfer is far from automatic even within a closely related continuum.

Interpreting these results in light of Pomak’s documented micro-variation, the baseline scores are consistent with a distribution-shift account: the Greece-variety training data may under-represent (or encode differently) constructions that are frequent in the Turkey variety, including contact-shaped argument-marking strategies, function-word inventories, and morphosyntactic exponence patterns. In low-resource settings, such shifts are amplified because the model has limited opportunity to learn variety-invariant abstractions, and instead relies heavily on surface cues whose realizations differ across dialects. The baseline therefore provides direct quantitative evidence that the existing UD resource does not straightforwardly ``cover'' the Turkey variety, motivating the dialect-matched mini corpus and the adaptation experiments presented in the following section.

\begin{table}[t]
  \centering
  \footnotesize
  \setlength{\tabcolsep}{4pt}
  \resizebox{\columnwidth}{!}{%
    \begin{tabular}{@{}lcc@{}}
      \toprule
      \textbf{Setup} & \textbf{Metric} & \textbf{Score} \\
      \midrule
      Train: Dataset A; Test: Dataset C & UAS & 59\% \\
      Train: Dataset A; Test: Dataset C & LAS & 51\% \\
      \bottomrule
    \end{tabular}%
  }
  \caption{Baseline cross-variety parsing performance. Dataset A is the existing Pomak UD treebank (primarily Greece variety), and Dataset C is the held-out Turkey-variety test set from a speaker not represented in Dataset B.}
  \label{tab:uas_las}
\end{table}

\subsection{Turkey-Variety Mini Corpus and Transfer Results}\label{sec:mini-results}

Recall that in order to assess how much \emph{dialect-matched} supervision can mitigate cross-variety degradation, and whether transfer-based training can further improve generalization under extreme data scarcity, we compare three training regimes. The first regime is the cross-variety baseline: the model is trained on the existing Pomak UD resource, which primarily reflects the Greece variety, and evaluated on Turkey-variety Pomak. The second regime is an in-variety condition in which the model is trained exclusively on our newly compiled Turkey-variety mini corpus. The third regime introduces transfer/adaptation: we leverage the broader syntactic coverage available in the larger UD dataset while anchoring the model in the Turkey-variety distribution via fine-tuning on the mini corpus.

Table~\ref{tab:scores_basic} compares the cross-variety baseline against the in-variety (mini-corpus-only) condition. Despite being dialect-matched, the mini-corpus-only model attains 50\% UAS and 44\% LAS, underperforming the model trained on the larger Greece-variety UD resource (59\% UAS / 51\% LAS). This outcome highlights the severity of the data-scarcity regime: 650 sentences do not provide sufficient coverage for the model to learn robust syntactic generalizations, particularly for lower-frequency constructions and long-distance dependencies. Importantly, this result separates two distinct constraints. Dialect match reduces distribution shift, but it cannot compensate for the statistical advantages of substantially larger supervision. In other words, within-variety training at this scale remains bottlenecked by coverage rather than by dialect mismatch alone.

\begin{table}[t]
  \centering
  \footnotesize
  \setlength{\tabcolsep}{3pt}
  \renewcommand{\arraystretch}{1.05}
  \begin{tabularx}{\columnwidth}{@{}Xcc@{}}
    \toprule
    \textbf{Training / Evaluation Setup} & \textbf{UAS} & \textbf{LAS} \\
    \midrule
    Dataset A $\rightarrow$ Dataset C & 59\% & 51\% \\
    Dataset B-train $\rightarrow$ Dataset C & 50\% & 44\% \\
    \bottomrule
  \end{tabularx}
  \caption{Cross-variety baseline versus in-variety training. Dataset A is the existing Pomak UD treebank (primarily Greece variety). Dataset B-train and B-dev are splits of our Turkey-variety corpus, and Dataset C is the held-out Turkey-variety test set from a different speaker.}
  \label{tab:scores_basic}
\end{table}

We therefore evaluate a transfer/adaptation regime that combines the strengths of both resources: the larger UD dataset supplies broader syntactic coverage and reduces estimation variance, while the mini corpus provides dialect-specific calibration for the target variety. Intuitively, the transfer setup encourages the model to learn higher-level abstractions from the larger resource and then adjust the mapping from surface cues to syntactic structure in the Turkey variety during fine-tuning. Under this regime, performance improves to 68\% UAS and 62\% LAS on a held-out Turkey-variety test set of 90 sentences (Table~\ref{tab:scores_all}). The size of this gain relative to both baselines suggests that additional supervision can be exploited without sacrificing dialect specificity, provided that training explicitly incorporates an adaptation step. Notably, the improvement is observed in both UAS and LAS, indicating that adaptation helps not only with head selection but also with relation labeling—components that are plausibly sensitive to contact-shaped differences in functional marking and argument encoding.

Overall, the results support a practical methodological conclusion for endangered languages with strong dialectal structure: when dialect-matched corpora are necessarily small, the most effective strategy is often not to train from scratch on the target dialect, but to treat dialect-matched data as an \emph{adaptation resource} integrated into a broader transfer pipeline. In our case, the transfer regime substantially improves parsing quality on the Turkey variety, suggesting a path forward for building usable syntactic tools even when in-variety annotation is limited.

\begin{table}[t]
  \centering
  \footnotesize
  \setlength{\tabcolsep}{3pt}
  \renewcommand{\arraystretch}{1.05}
  \begin{tabularx}{\columnwidth}{@{}Xcc@{}}
    \toprule
    \textbf{Model / Training Regime} & \textbf{UAS} & \textbf{LAS} \\
    \midrule
    Original Pomak UD (Greece variety) & 59\% & 51\% \\
    Mini corpus & 50\% & 44\% \\
    Transfer / adaptation (A$\rightarrow$B) & 68\% & 62\% \\
    \bottomrule
  \end{tabularx}
  \caption{Parsing performance under three regimes: out-of-variety baseline, in-variety mini-corpus-only training, and transfer/adaptation leveraging both resources.}
  \label{tab:scores_all}
\end{table}

\section{Discussion}\label{sec:discussion}

Our results provide converging evidence that Pomak NLP must be treated as a \emph{dialect-sensitive} low-resource problem rather than a single homogeneous target. Training on the existing Pomak UD resource (primarily Greece variety) yields only moderate cross-variety generalization to the Turkey variety (59 UAS / 51 LAS), indicating a substantial distributional mismatch despite genealogical relatedness. This pattern is consistent with the linguistic literature portraying Pomak as a contact-shaped dialect continuum: sustained bilingualism and long-term contact with Turkish and Greek can induce both lexical turnover and systematic morphosyntactic reorganization, including shifts in functional inventories and argument-encoding strategies as also shown in \citep{karakas2022}. In such circumstances, models trained on one variety may internalize variety-specific correlates of syntactic relations---for example, the co-occurrence patterns between function words, case/marker realizations, and dependency labels---that do not transfer faithfully to another variety, yielding errors in both attachment (UAS) and relation assignment (LAS). The non-trivial UAS--LAS gap in the baseline further suggests that relation labeling is particularly sensitive to these shifts, plausibly because UD relation inventories are mapped from surface cues that may change under contact (e.g., oblique marking strategies, adpositional elements, or the distribution of possessive morphology).

At the same time, the comparison between in-variety and transfer-based training shows that \emph{dialect match alone is not sufficient} when labeled supervision is extremely limited, echoing core results in domain and parser adaptation that low-resource target settings typically benefit from transfer and semi-/self-supervised adaptation \citep{daume-iii-2007-frustratingly,jiang-zhai-2007-instance,mcclosky-etal-2006-reranking,gururangan-etal-2020-dont}. Training exclusively on the 650-sentence Turkey-variety mini corpus leads to lower performance (50 UAS / 44 LAS) than the cross-variety baseline, despite being distributionally better aligned with the test variety. This outcome may suggest a central tension for endangered-language NLP, which is that the very scenarios that most demand dialect-matched resources are those where annotation budgets are smallest, and small in-variety corpora may lack the coverage required to estimate robust syntactic generalizations. In practice, the model must generalize across sparse constructions, long-distance dependencies, and low-frequency lexical types; with only a few hundred sentences, learning is dominated by estimation variance and incomplete coverage rather than by mismatch alone. This result also suggests that, for Pomak specifically, dialect-conditioned lexical/functional variability may interact with orthographic non-standardization to inflate the effective type inventory, further increasing sparsity under small corpora.

Crucially, the transfer/adaptation regime resolves this tension by combining broad-coverage supervision from the larger UD resource with dialect-specific calibration from the mini corpus. Fine-tuning/adaptation improves performance to 68 UAS / 62 LAS on the Turkey-variety test set, yielding gains over both baselines. We interpret this improvement as evidence that (i) the larger resource supplies essential syntactic coverage and stabilizes parameter estimates, while (ii) the dialect-matched mini corpus effectively ``reanchors'' the learned representations in the target variety, adjusting the mapping from surface diagnostics to syntactic structure. That the improvement appears in both UAS and LAS is particularly important: adaptation helps not only with head selection but also with relation labeling, implying that the fine-tuning signal corrects systematic mismatches in how functional material and argument marking correlate with UD relations in the Turkey variety. Methodologically, these findings align with prior work showing that multi-treebank/multilingual supervision provides broad syntactic coverage for low-resource parsing and that an additional fine-tuning/adaptation step on limited target-variety data can substantially improve both unlabeled and labeled attachment, effectively calibrating the parser to the target distribution \citep{stymne-etal-2018-parser,kondratyuk-straka-2019-75,stymne-2020-cross,glavas-vulic-2021-climbing,velldal-etal-2017-joint,mokh-etal-2024-domain}.

The results also have implications for resource design. First, they argue for explicitly encoding \emph{variety metadata} in Pomak resources (e.g., dialect tags, speaker region, orthographic conventions), since the performance gap demonstrates that pooling across varieties without control can obscure distribution shifts that matter for modeling. Second, they motivate targeted annotation strategies that prioritize constructions likely to differ across varieties---for example, oblique marking and ditransitives, functional elements like \textit{na}-type markers, and contact-sensitive possessive and case morphology. In future work, this could be operationalized via active learning criteria that are explicitly variety-aware (e.g., uncertainty sampling over dialect-diagnostic contexts), thereby increasing the return on limited annotation effort. Finally, while our experiments focus on dependency parsing, the same logic extends to downstream tasks (information extraction, morphology, ASR/MT): contact-conditioned dialect differences can create systematic error modes that are not reducible to ``noise'' and therefore benefit from explicit adaptation.

Several limitations point to concrete next steps. The Turkey-variety corpus remains small, and our evaluation set should be expanded to reduce uncertainty and enable more fine-grained error analyses by construction type. In addition, orthographic and tokenization choices are likely to interact with character-level modeling; systematically comparing normalization schemes and subword segmentation strategies may further improve robustness. Finally, more diagnostic analyses---e.g., confusion matrices over POS and dependency labels, and targeted evaluation on constructions known to differentiate varieties---would sharpen the linguistic interpretation of the observed gains and help identify which aspects of contact-shaped variation are most consequential for syntactic modeling.

\section{Conclusion}\label{sec:conclusion}

This paper investigated whether Pomak’s contact-shaped dialectal variation poses a measurable challenge for syntactic processing and whether transfer-based modeling can mitigate this challenge under severe data scarcity. Using the existing Pomak UD resource (primarily representing the Greece variety), we found only moderate cross-variety generalization to the Turkey variety, consistent with a substantial distribution shift in the surface cues that support both attachment and relation labeling. To directly probe this mismatch, we constructed a Turkey-variety mini corpus of 650 sentences compiled from six native speakers and curated thesis data, yielding UD-compatible supervision explicitly targeting the Uzunköprü variety.

Our experiments show that dialect match alone is not sufficient at this scale: training exclusively on the mini corpus underperforms training on the larger out-of-variety UD resource, highlighting the limits of purely in-variety supervision in the extreme low-resource regime. By contrast, a transfer/adaptation strategy that combines broad-coverage supervision from the larger UD resource with fine-tuning on the dialect-matched mini corpus yields the strongest results, improving both UAS and LAS on Turkey-variety evaluation. Taken together, these findings support a practical methodological recommendation for endangered, dialectally diverse languages: when dialect-specific corpora are necessarily small, they are most effective when used as an adaptation signal integrated into a broader transfer pipeline, rather than as a stand-alone training resource.

\section{Limitations and Future Directions}\label{sec:limitations}

Several limitations of the present study point to clear directions for future work. First, the current analysis does not yet isolate parser behavior on dialect-sensitive constructions that are especially relevant for Pomak, such as oblique-marking strategies, ditransitives, case-marking asymmetries, or function-word realizations shaped by contact. As a result, while the quantitative comparison demonstrates that cross-variety transfer is imperfect and that fine-tuning improves performance, the present paper cannot yet determine which specific grammatical phenomena are driving the observed degradation and subsequent gains. Besides, the Turkey-variety corpus remains small, and this necessarily constrains both modeling and evaluation. Although the 650-sentence corpus is sufficient to demonstrate the value of dialect-matched adaptation, it is too limited to support fine-grained analysis across a broad range of constructions, genres, and speaker profiles. Future work should therefore expand the Turkey-variety dataset substantially, both by increasing the number of annotated sentences and by broadening the range of speakers and usage contexts represented in the corpus. 

\section{Acknowledgements}

I am deeply grateful to Balk{\i}z {\"O}zt{\"u}rk for her guidance on endangered languages. I also thank Emirhan {\"O}zkan for his help in reaching the Pomak language community. Finally, I sincerely thank the Pomak speakers who generously shared their language with me; this work would not have been possible without their time, trust, and knowledge.

\section{Bibliographical References}\label{sec:reference}
\bibliographystyle{lrec2026-natbib}
\bibliography{lrec2026-example}

\clearpage
\appendix

\end{document}